OPEN

# Modelling the long-term fairness dynamics of data-driven targeted help on job seekers

Sebastian Scher[1]✉, Simone Kopeinik[1], Andreas Trügler[1,2,3] & Dominik Kowald[1,2]

**The use of data-driven decision support by public agencies is becoming more widespread and already influences the allocation of public resources. This raises ethical concerns, as it has adversely affected minorities and historically discriminated groups. In this paper, we use an approach that combines statistics and data-driven approaches with dynamical modeling to assess long-term fairness effects of labor market interventions. Specifically, we develop and use a model to investigate the impact of decisions caused by a public employment authority that selectively supports job-seekers through targeted help. The selection of who receives what help is based on a data-driven intervention model that estimates an individual's chances of finding a job in a timely manner and rests upon data that describes a population in which skills relevant to the labor market are unevenly distributed between two groups (e.g., males and females). The intervention model has incomplete access to the individual's actual skills and can augment this with knowledge of the individual's group affiliation, thus using a protected attribute to increase predictive accuracy. We assess this intervention model's dynamics—especially fairness-related issues and trade-offs between different fairness goals- over time and compare it to an intervention model that does not use group affiliation as a predictive feature. We conclude that in order to quantify the trade-off correctly and to assess the long-term fairness effects of such a system in the real-world, careful modeling of the surrounding labor market is indispensable.**

Data-driven methods for decision support—also known as data-informed decision support systems, AI-based decision support, or algorithmic decision-making—form useful technologies in many fields and get more and more widespread, not only in the private but also in the public sector[1,2]. However, this also raises concerns among the general public, as AI-based systems are prone to replicate biases present in data and application design. For instance[3], found that users attributed females receive a lower number of high-paid job-adds than similar male users. While the data used may be correct in its collection and historical representation, it often depicts outdated societal norms and values, capturing historical inequities and cultural biases[4]. When entering data-driven applications, the resulting discrimination adversely affects minorities and groups that have already been discriminated against and disadvantaged in the past and consequently creates a reinforcement loop[5]. Some applications—for example, social scoring by authorities—are considered so problematic that in the European Union there is a plan to prohibit them [6].

The labor market is an area in which the use of AI-based decision support must be examined with particular care, as there is a long tradition of discrimination against social groups in the labor market, for instance, based on ethnicity[7] or gender[8], which in turn is reflected in data.

Public employment services (PES) support people in finding jobs and play a very important social role in many countries. In the European Union, access to free employment services is even a fundamental right (art. 29 EU fundamental rights charter). Recently, the Austrian Public Employment Service (AMS) has started to use an AI-based system that categorizes job-seekers into three groups according to individuals' low, moderate, or high prospects in the labor market. This categorization allows providing different types (qualities) of help depending on the individual's group affiliation. Also, predominantly supporting the group with moderately good prospects is considered most (cost) efficient, as their labor market prospects can be raised to an acceptable level with relatively little effort. For individuals in the low-prospect group, on the other hand, more effort would be needed to achieve the same outcome.

In the AMS system, prospects are informed by calculated probabilities (i.e., predictions) of people finding a job in the near future (i.e., the next couple of months)[9]. The prediction model is trained on employment data collected in the past and therefore exhibits a historical bias. Socio-cultural norms and prejudices are reflected and cause attributes of a person, such as gender and caretaking obligations, to be most influential on the predictive outcome. The fact that gender is used as a predictor, as well as sociopolitical issues related to the "efficient" distribution of public services, led to a broad public debate about the ethical implications of using the system[10–12].

[1]Know-Center GmbH, Graz 8010, Austria. [2]Institute of Interactive Systems and Data Science, Graz University of Technology, Graz 8010, Austria. [3]Department of Geography and Regional Science, University of Graz, Graz 8010, Austria. ✉email: sscher@know-center.at

The main political goal behind systems such as the AMS system is usually the wish to make public services—in this case employment services—more efficient. However, this could potentially have harmful and unintended consequences. Existing group inequalities could be reinforced by the systems interventions over time, or new inequalities might emerge. However, if and to what extent this happens in a particular setting is not easy to predict. Dynamic systems can often act in unintuitive ways, and the inclusion of regularly updated statistical models makes the system even harder to understand. It has been shown that a system that is fair in a static context can still produce unfair results in the long run if it is regularly updated and provides feedback to the environment[13]. Other results show that the situation of a system that is unfair in the beginning might also worsen if a specific privacy method is applied [14]. Despite these difficulties, non-discrimination and fairness are among the ethical and legal requirements for AI systems[15,16], and it is thus essential to find ways to assess fairness aspects also of labor market intervention systems that use statistics and AI. A known issue is that the stakeholders involved in the development cycle, even if sensitized to the concern, often lack experience, processes, and tools to manage the complex set of issues[17,18]. Our paper is inspired by the ideas and problems of the AMS system, but it is not a study of this particular system. Instead, we focus on the general idea of such systems and their long-term effects.

This paper serves two main purposes: (1) we provide an introduction to the complexity of assessing the long-term fairness effects on the population if a public authority provides targeted help in the labor market, based on data-driven methods that include protected attributes (e.g., gender). Targeted help, for the purposes of this paper, means that individuals or groups of individuals receiving help are selectively chosen based on predetermined criteria. (2) We provide an answer to the question "How can we assess long-term fairness in a dynamical system such as a labor market?" For this, we develop and present an approach on how to actually assess such long-term fairness impacts in a dynamical system such as a labor market.

Additionally, our study means to highlight the benefits of quantitative modeling of the surrounding environment as an essential part of the assessment of causal long-term effects. We focus on the following main aspects:

- Trade-offs between different long-term fairness goals (e.g., reducing inequality between groups versus correctly assessing individuals' labor market prospects) when a PES provides targeted aid to job-seekers.
- Impact of targeted aid- versus non-targeted aid- on the long-term fairness of public authority interventions.

To investigate these aspects, we propose a combination of dynamical numerical modeling and data-driven models. The model captures the principle dynamics of a labor market situation in which a public authority intervenes with targeted aid. It consists of a replenishing pool of job seekers, defined via a skill model, the labor market, in which job-seekers do or do not get jobs, and the PES, which intervenes and changes the skills of the job-seekers. All these parts are abstractions of the real world and are kept as simple as possible while still capturing the basic dynamics.

The individual skill model is defined as simply as possible while at the same time being sophisticated enough to account for inequalities and, optionally, either full or incomplete knowledge of an individual's skills. Due to the lack of openly available empirical data, we use synthetic data in our study.

We assume a population of individuals, where the prospect on the labor market of each individual is controlled by the personal skill set of that individual, described by a set of independent skill features. Additionally, each individual belongs to one of two groups, described by a protected attribute (e.g., gender). The average skill level between the two groups is not the same but shows significant overlap. If an observer has knowledge about all skill features of an individual, they can accurately compute the total skill level of that individual, which means that knowledge about the protected attribute (i.e., which group the individual belongs to) does not yield any additional information with regard to the individual's skill level.

We further assume that there is a public authority that helps individuals in improving their skills in the labor market, which we will call the Public Employment Service (PES). To support the improvement of the skills of an individual, the PES provides access to services that are selected according to the individual's current prospects in the labor market. While the labor market (employers) has access to all skill features, the public authority, however, does not have access to all skill features but only to a subset. This assumption can be justified by the fact that in real life, while both employers and public authority will have some common information about the job-seekers (degrees, years of work experience, etc), employers will have both more resources and more branch-specific knowledge for evaluating applicants (e.g., exact degrees, specializations, soft skills, etc). In addition, it has knowledge about the group affiliation (i.e., protected attribute). Because the total skills are not evenly distributed across the two groups, the knowledge about an individual's group affiliation combined with historical data gives probabilistic information on their real skills: if the individual belongs to the group that has on average higher skills, also the likelihood that this individual has high skills is larger than if it belongs to the other group, even if there is no other distinction in attributes. This concept has previously been discussed in economic research[19]. Using this additional information, however, has two potential problems: (i) the model is probabilistic, and thus, the resulting predictions are only accurate *on average*, and (ii) the model is based on a protected attribute—therefore, legal and/or ethical reasons might prohibit utilizing this information, as it results in different treatment solely based on the affiliation to a certain group given by the protected attribute.

In order to gain a better understanding of the implications different approaches might entail, we compare two prediction models that the PES could implement: one that uses the protected attribute and one that does not. With this, we can study the trade-off between mitigating disparities in personal skills among the two groups and the aim of preventing misclassifications based on a protected attribute.

Additionally, we make assumptions about the labor market and encode these assumptions in a simple dynamical model. The model has a pool of job-seekers with an influx and an outflux. The PES provides targeted help to the current job-seekers. The targeting of the help is based on the skills of individuals and the historical model record of how long it took individuals with different skills to find a job. With this model, we consider different scenarios/approaches of the PES on how it distributes its limited resources across individuals with different assumed skills. Furthermore, we test different assumptions about the job market (e.g., biased and unbiased) and investigate how it affects the impact of targeted vs. non-targeted help. The model we develop and use in this study is "dynamic" on two different levels: first, as it is an agent-based model, it is dynamic at the level of individuals; second, as a consequence of the intervention of the PES, it is dynamic in the adaptation of average skill levels. The latter is similar to what in the economics literature would be referred to as transition paths between different policy states.

## Related work

From a fundamental rights perspective, the issue of fairness in data-driven applications is discussed in several reports of the European Union Agency for Fundamental Rights (FRA)[20–22]. There are different definitions of fairness, depending on context, application and world-view, which occasionally contradict each other. From a legal perspective, a major problem is that court decisions are highly tailored to specific circumstances, which contrasts with quantitative, generic measures of fairness[23]. A wide variety of quantitative fairness metrics and debiasing algorithms have also been proposed in research, e.g.,[24,25],[26] investigates what people perceive as most fair and find that demographic parity (a.k.a. statistical parity) receives the highest level of agreement in several cases presented. They argue against the common practice of optimizing AI-based decisions toward multiple fairness goals, but to select the most meaningful metric in terms of social context. In this paper, we take as a sample environment - and thus social context - the labor market and the long-term effects of tailored measures to support job seekers. This is inspired by the AI-based decision support system used by the Austrian PES (AMS) that caused a great level of public controversy and already has been the subject of previous studies. Lopez[10] for instance, elaborates extensively on algorithmic details, as well as their underlying human-based decisions and their possible implications on the affected population. An emphasis is set on gender aspects and potential (intersectional) discrimination. Authors also remark on the lack of research with respect to whether and how to include gender as an attribute in such an algorithm. Allhutter et al.[11] takes an approach based on critical studies and fairness to discuss the "inherent politics of the AMS algorithm". While in our paper we do not explicitly investigate the AMS algorithm, we contribute to this line of research by investigating the long-term impact of different intervention models on job seekers and, in particular, by exploring the algorithmic consideration of a protected attribute that distinguishes groups (e.g., gender). To this end, we introduce a dynamical modeling approach that complements previous research on long-term dynamics of fairness.

**Long term dynamics of fairness.** Liu et al.[27] introduces a formal, one-step feedback model to estimate the long-term impact of fairness constraints. It is presented by the example of a credit distribution scenario, but can, however, also be adapted to other domains given necessary domain knowledge. Mouzannar et al.[28] goes beyond this approach and introduces a formal, yet flexible model that allows the study of both economic utility and social equality as a consequence of fairness interventions. In our model, we follow a more dynamic modeling approach and investigate fairness according to other characteristics, going beyond the change in population mean, while our study is also more specific to labor market interventions. Instead of adopting a general approach, we develop a specific dynamical model that reflects our problem setting. Kannan et al.[29] discusses fairness in colleague admission and graduate hiring based on a two-stage model. The study concludes that under real-life conditions, two defined fairness goals (i.e., being admitted and hired independent of group membership) are unreasonable to achieve. The study clearly extends simple static models, however, does not intend to study the long-term impact of the fairness goals but rather aims to identify environmental variables that could allow the achievement of defined fairness goals. D'Amour et al.[13] presents an extensive aspect in how results of long-term modelling may differ from static evaluation settings. In three simulation scenarios (i.e., loan allocation, college admission, and attention allocation) they show how simple agent-environment models evolve over time. Their results highlight the need to assess the fairness of algorithmic systems in continuous time steps.

We add to this line of research and introduce a more complex dynamic modeling approach that depicts a rather systemic viewpoint on data-driven decision-making implications. This results in more complex, quantitatively evaluated simulations that, however, still simplify the real-world setting. We also study the dilemma between individual and group fairness that has been commonly discussed in AI applications, particularly in regard to data-driven decision support[30].

**Economics.** We use the concept of "labor-market-models" in a way that is targeted toward the main aspects of our study. In economics, a number of different labor market models are used to study the supply and demand of labor (e.g.,[31]). In that context, our modeling approach can be seen as agent-based modeling, which is also widely adopted in economic research (see[32] for a historical overview of agent-based modeling of labor markets). Chaturvedi et al.[33] builds an agent-based model of the labor market for research purposes, in which agents are individual persons. Discrimination in the labor market is a widely studied topic in economics. Seminal work was done by Ken Arrow, who studied discrimination in the labor market back in the 70s[19]. One explanation for discriminatory results Arrow gives is imperfect information, which we consider a variable in our model. Caine[34] gives and overview on the early work on labor market discrimination. In[35], the authors give an over-view of theory and empirical evidence of racial discrimination in the labor market. A recent text-book on the topic is[36] (especially chapter 12). Finally[37], argues that individual fairness constraints are insufficient to remove racial inequality from the US labor market. They suggest a "dual labor market" that could solve this problem by applying a dynamical approach. They also argue that such further-reaching approaches will be more and more important if employment processes are continuously automatized. Similar to our work and to[13], they abandon the concept of static fairness. However, our work focuses on the trade-offs between different long-term fairness goals, as described in the subsequent sections. Cohen et al.[38] studies the efficiency of recruiting practices from an employer's point of view, including the incorporation of fairness constraints. The unique point of our study is the inclusion of a PES that uses a continuously updated data-driven decision model. To the best of our knowledge, this has not been done before.

## Methods

**Personal skill model and data generation.** In our personal skill model, each individual has a personal skill set $s_{real}$ that is composed of two independent skill features $x_1$ and $x_2$.

$$s_{real}(x_1, x_2) \equiv \frac{1}{2}(x_1 + x_2) \tag{1}$$

We call it "real" because we will differentiate it from observed and from predicted/assumed skills later on. In addition, each individual has a binary protected attribute $x_{pr}$ that can have values of 0 and 1. In reality, this could for example be female or male, but here it is used in an abstract way. Central here is that the definition of $s_{real}$ does not explicitly contain $x_{pr}$.

We draw $x_1$ and $x_2$ from uncorrelated truncated normal distributions. This means there is no correlation between $x_1$ and $x_2$. We use normal distributions because personal features such as "talent" are usually assumed to be normally distributed (e.g.,[39]). Truncation is used to ensure that no one has $x_1$ and/or $x_2$ higher than the maximum reachable values in the intervention model (see Section "Intervention model" below). The distribution is truncated at plus-minus two times the standard deviation. Therefore, despite the truncation, the distribution is very similar to a non-truncated normal distribution, and thus appropriate for describing personal features. Furthermore, we assume that $x_1$ is completely independent of $x_{pr}$:

$$x_1 \in \mathcal{N}_{trunc}(0, 1), \tag{2}$$

The values of the binary protected attribute have equal probability:

$$x_{pr} \in \{0, 1\}, \quad p\left(x_{pr} = 0\right) = p\left(x_{pr} = 1\right) = \frac{1}{2} \tag{3}$$

In words, for generating our artificial population, we draw $x_1$ from a truncated normal distribution (truncated at $x_{max}$), and $x_{pr}$ from the binary distribution $\{0, 1\}$ with uniform probability.

Thus, the probability of an individual belonging to a particular group (with respect to the protected attribute) is 50% for both groups, and both groups are therefore of equal or near equal size.

The second skill feature, $x_2$, is correlated with $x_{pr}$ and is generated with the following formula:

$$x_2 \in \frac{1}{2}\left\{\alpha_{pr} \cdot \left(x_{pr} - \frac{1}{2}\right) + \mathcal{N}_{trunc}(0, 1)\right\} \tag{4}$$

The parameter $\alpha_{pr}$ controls how much $x_2$ is higher on average in the privileged group compared to the underprivileged group. The factor $\frac{1}{2}$ is subtracted from $x_{pr}$ to ensure that $x_2$ has a mean of zero. When $x_2$ is generated this way, the individuals with $x_{pr} = 0$ have *on average* lower $x_2$, and therefore on average lower $s_{real}$. To reflect this, we will from now on call the group of individuals with $x_{pr} = 0$ the *underprivileged group*, and individuals with $x_{pr} = 1$ the *privileged group*. Importantly, however, not all individuals in the underprivileged group have low $x_2$ and low $s_{real}$. There are individuals in the privileged group that have lower skills than some individuals in the underprivileged group, and there are individuals in the underprivileged group that have a skill that is above the population mean. The joint distribution of $x_1$ and $x_2$ and the distribution of $s_{real}$ of a sample from the background population is shown in Fig. 1.

From the way $x_1$ and $x_2$ are generated and the fact that $s_{real}$ per definition (Eq. 1)) can be completely inferred from $x_1$ and $x_2$, follow two central facts: Given $x_1$ and $x_2$, there is no additional information contained in $x_{pr}$ when one wants to infer $s_{real}$ (even though $s_{real}$ is correlated with $x_{pr}$). If, however, one has only access to $x_1$ and at the same time, information on the distribution of $s_{real}$ over the two groups (e.g., the mean of $s_{real}$ separately for each group), and one wants to infer $s_{real}$, then including $x_{pr}$ in addition to $x_1$ in a statistical prediction system yields additional information, even though $s_{real}$ is completely defined by $x_1$ and $x_2$. This will form the backbone of our study. If it is for example known that an individual has an average value of $x_1$, but we do not know $x_2$, then knowing $x_{pr}$ will be decisive in estimating whether $s_{real}$ of that person is below or above average: if the individual belongs to the underprivileged group, then the expectation would be that $s_{real}$ is below average, but if the individual, with an unchanged value of $x_1$, is in the privileged group, then the expectation would be that $s_{real}$ is above average.

For our model, we assume that there is an (unlimited) *background-population* pool with the distribution of $x_1, x_2, x_{pr}$ and $s_{real}$ described by Eqs. (1)–(4). This background population and the distribution of the features of the individuals do not change throughout a model run, but acts as a pool for refilling the pool of job-seekers.

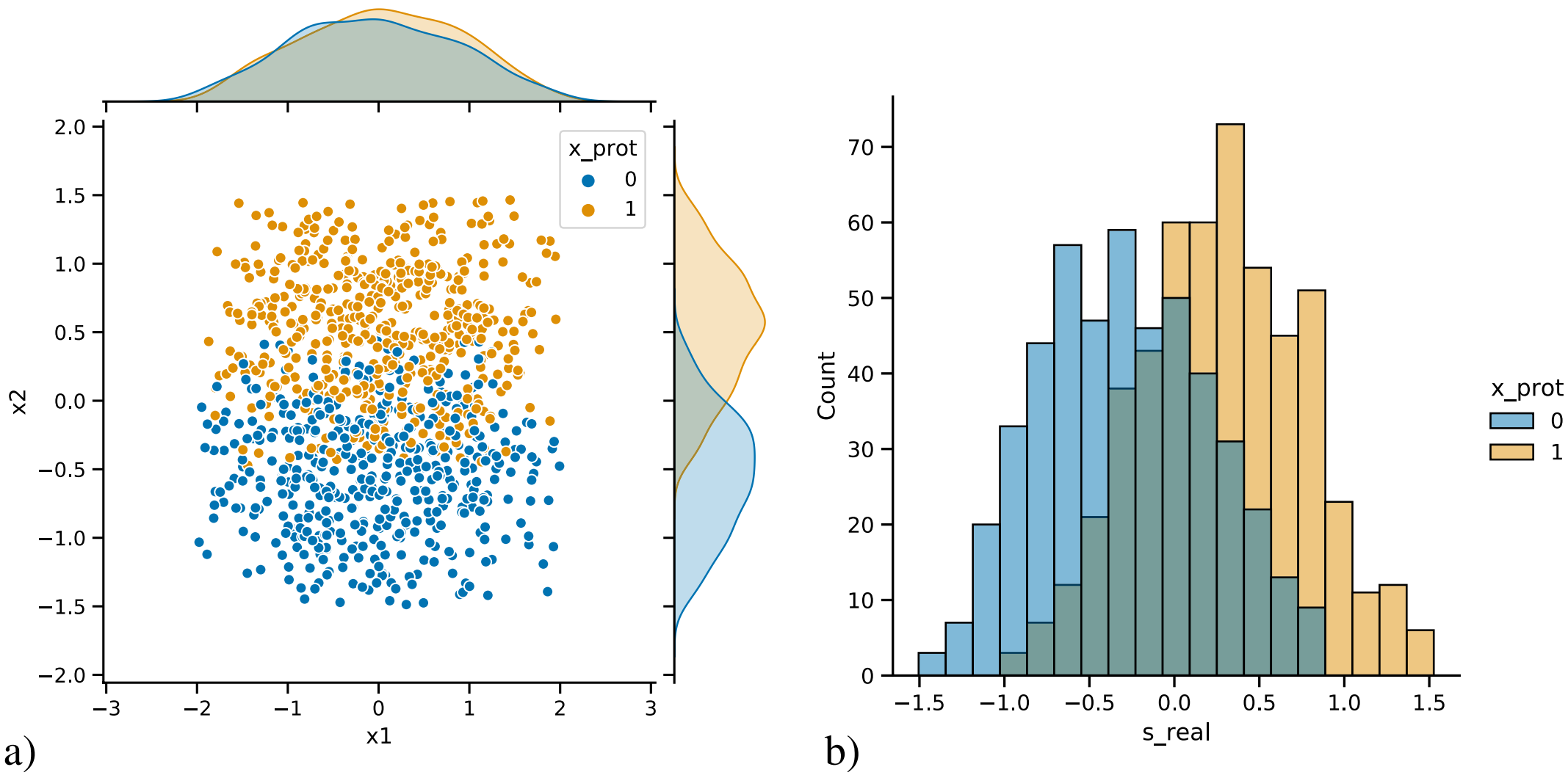

**Figure 1.** Sample of the background-population (**a**) Distribution of the two skill features, (**b**) Distribution of total skills ($s_{real}$), both split up according to the binary protected attribute. In (**b**), both colors are half-transparent, and the overlapping region is therefore depicted by the mixed color.

**Prediction model.** The prediction model is used by the PES in order to group individuals into a high-prospect and a low-prospect group, depending on the expected time-span $T_u$ they will be unemployed (high-prospect = expected to find a job seen without help, low-prospect = expected to take longer to find a job without help). This is based on the time it took individuals to find a job in the past. The history is continuously build up throughout a model run with all individuals that found a job. The basis of this study is that the PES has access to an incomplete set of skill features only, namely solely to $x_1$, and additionally access to $x_{pr}$. This assumption is reasonable because, in the real world, the PES will only have limited information about an individual (e.g. their education level and employment history), without access to more detailed information such as detailed CVs, job interviews, tests, etc. The simplest way to model this is through having 2 skill features, of which the PES can observe only one. To estimate (predict) the prospect group (above or below average $s_{real}$) from this, logistic regression is used to create the main (full) prediction model:

$$P_{full}\left(T_u > T_u^{\gamma}|x_1, x_{pr}\right) = \frac{1}{1 + e^{-(\alpha_1 x_1 + \alpha_2 x_{pr} + \beta)}} \tag{5}$$

Here, the parameters $\alpha_1, \alpha_2$ and $\beta$ are estimated from the historical record, and $T_u^{\gamma}$ is the threshold set on the unemployment time $T_u$ for dividing the low and the high prospect group.

Additionally, we use a second prediction model, which we will call the *base model*, that does not use $x_{pr}$:

$$P_{base}\left(T_u > T_u^{\gamma}|x_1\right) = \frac{1}{1 + e^{-(\alpha^* x_1 + \beta^*)}} \tag{6}$$

with free parameters $\alpha^*$ and $\beta^*$ fitted on the historical record. Logistic regression falls in a broader category of methods often referred to as supervised machine learning. Also, other supervised machine-learning algorithms, such as neural networks, could be used in the prediction model. Our choice for logistic regression was motivated by the fact that (i) it is the same method as used in the model that inspired our work i.e., the AMS-system, and (ii) it is a simple and easy to interpret method which allows for a certain level of transparency (in contrast to e.g., neural networks). A comparison of the full and the base model is necessary for investigating whether there are differences between different long-term fairness goals, as using the full prediction model conflicts with the fairness goal of not using the protected attribute but might be out-weighted by other long-term effects.

**Labor market model.** The labor market is modeled via a probabilistic function that for each individual defines the probability of finding a job at the current timestep, where this probability depends on $s_{real}$ of that individual. We model the dependence on $s_{real}$ as a logistic function:

$$P\left(job|s_{real}\right) = \frac{1}{1 + e^{-(\alpha_l s_{real} - \beta_l + b)}} \tag{7}$$

where $\alpha_l$ and $\beta_l$ are fixed parameters. The parameter $\alpha_l$ controls the influence of $s_{real}$ on the probability of finding a job, $\beta_l$ sets at which value of $s_{real} + b$ the probability is 0.5. At each timestep, $P\left(job|s_{real}\right)$ is computed for each individual within the pool of job-seekers. Each individual is removed from the pool of job seekers with a probability of $P$. The value $b$ describes how biased the labor market is in favor of the privileged group. It is computed from a fixed labor market bias parameter $\beta_b$ and the protected attribute:

$$b = \beta_b(x_{pr} - 0.5) \tag{8}$$

With $\beta_b = 0$ the labor market is unbiased, with $\beta_b > 0$ it is biased in favor of the privileged group. We use two different values in our experiments: 0 (“unbiased“), and 2 (“biased“).

The choice for a logistic labor market function was made because it satisfies the following intuition about the labor market: if an individual has very low skills, then the probability of finding a job is very low (close to zero), and if the skills slightly increase, then the chance is still very low. There is a soft threshold that one needs to reach in order to have a reasonable chance. Above this soft threshold, increases in $s_{real}$ have a strong impact. Thus, the higher skills an individual has, the higher the chances of finding a job. Eventually, however, this reaches a plateau, as the probability of finding a job is already close to 1, and additional skills do basically not change anything anymore. The parameters $\alpha_l$ and $\beta_l$ define this “middle” region, in which changes of $s_{real}$ have a strong impact on the probability. $\beta_l$ defines the position of this middle region, and $\alpha_l$ how broad/steep it is.

Note that we made $P(job|s_{real})$ independent of the time an individual is already unemployed. The intuition behind this is that in our idealized setting, the skills of an individual are solely defined by $s_{real}$, which the labor market knows. Therefore, in this setting, the fact that someone has been unemployed for a long time does not yield additional information about their skills. In reality, this might not necessarily be the case, since long-term unemployment as additional information could be a reason for an employer not to hire someone.

**Intervention model.** The intervention model describes the effect that the helping intervention of the PES has on the individual. The treatment of individuals differs between the high- and the low-prospect group in two ways: (i) in the amount of help (increase of $x_1$ and $x_2$) they receive, and (ii) in how long this help takes. The high-prospect group receives the help immediately and is available on the labor market in the next timestep. The low-prospect group, on the other hand, receives help that takes time and removes them from the labor market for a certain period of time $\Delta T_u$. This is a simplified version of the current strategy of the Austrian PES (AMS).

The change in the individual skills features $x_1$ and $x_2$ depends on the current values, with decreasing increments as the skill features grow, approaching the limits set by the constants $x_1^{max}$ and $x_2^{max}$:

$$x_1^{t+1} = max\left[x_1^t + k\left(x_1^{max} - x_1^t\right), x_1^t\right] \tag{9}$$

$$x_2^{t+1} = max\left[x_2^t + k\left(x_2^{max} - x_2^t\right), x_2^t\right] \tag{10}$$

The model parameter $k$ defines how fast $x_1$ and $x_2$ grow. For simplicity, we use the same growth rate for both skill features. The choices for the value of $k$ are described in Section “Scenarios”. Since individuals classified as low-prospect are removed from the active group for $\Delta T_u$ timesteps, their skills are updated $\Delta T_u + 1$ times (by applying Eqs. (9) and (10) $\Delta T_u + 1$ times) to account for that. Individuals that have been unemployed for too long (set by $T_u^{max}$) leave the system automatically.

As the model has random components, we run each simulation 10 times and average the results. All parameters of the model and their values are listed in Table B1 in the Supporting Information. The parameter values were set after testing different combinations. As we do not attempt to model an existing real-world setting, we have chosen a configuration that reaches reasonable equilibria for our main experiments, and the model with these parameters does not necessarily correspond to a specific use case. The sensitivity of the parameter choices is tested with additional experiments (see Supporting Information). An overview of the labor market and PES model is shown in Fig. 2.

**Intervention scenarios.** The value of $k$ in the intervention model from Eqs (9) and (10) is central to our study, as it defines how strongly the intervention model affects different people. Testing different values is necessary to determine if there are differences between targeted and non-targeted help. To this end, we make $k$ dependent both on the real prospect group $C_r$, and the prospect group $C_{pr}$ predicted by the prediction model. The fact that the growth rate is made depended on the predicted prospect group reflects the idea of targeted help for different prospect groups, and that a prediction model is used to do this. Our idea is not only that different prospect groups receive a different *quantity* of help, but also a different *quality* that is better suited for that prospect group. Therefore, we also make $k$ dependent on the real prospect group of each individual, as arguably if a specific type of help is better suited for the low then for the high prospect group, this will have the adverse effect for an incorrectly classified person.

The real prospect of a person cannot be precisely known, as the prospect is the *expected* time $T_u$ that the individual will be unemployed. The real-time that this individual will need cannot be known, as it evolves from the model and is linked to $s_{real}$ (but only in a *probabilistic* way). However, for the effects of the intervention model, we need something that is at least close to the real prospect. We, therefore, define the “real prospect” of an individual as $T_u$ estimated from the historical data, using $s_{real}$ as a predictor (in contrast to the predicted prospect, which uses $x_1$ and—if the full prediction model is used—$x_{prot}$ as predictor). As for the predicted prospect, high and low prospect groups are defined via the same threshold $T_u^\gamma$, and each group is predicted with logistic regression.

Since both the real prospect group and the predicted prospect group are binary, this leads to a 2*x*2 matrix $k_{ij}$:

| | **predicted low** | **predicted high** |
|---|---|---|
| real low | $k_{11}$ | $k_{12}$ |
| real high | $k_{21}$ | $k_{22}$ |

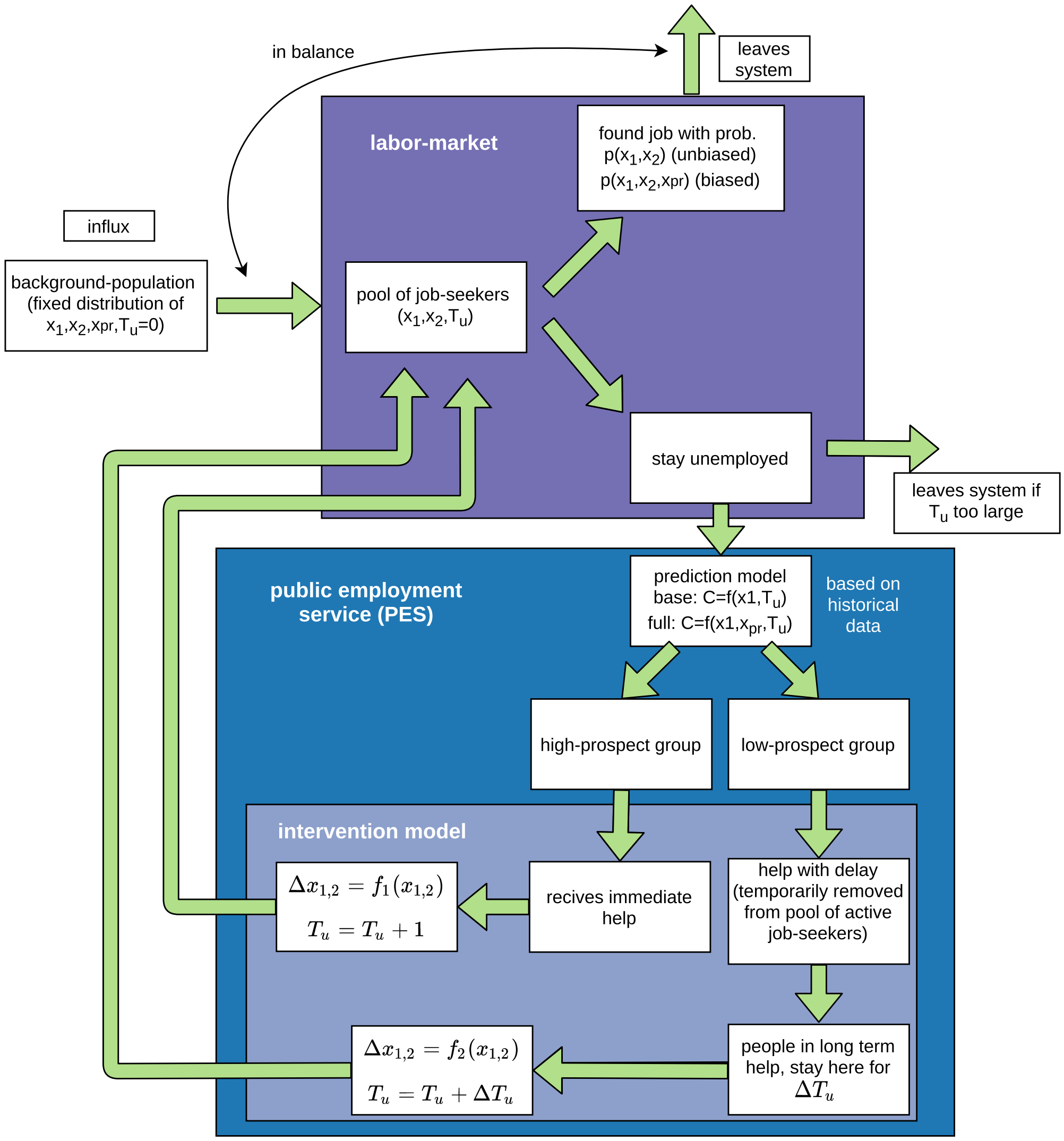

**Figure 2.** Outline of the labor market and PES model developed for this study. The labor market selects job-seekers from the pool of job seekers, with a probability dependent on the skills of the individual. Individuals who find a job leave the system, and individuals who have not found a job are transferred to the PES. The PES divides them into two groups, according to their predicted prospects in the labor market. The group with high prospects receives help (increase in skills) immediately and goes back to the pool of job seekers in the next timestep. The individuals in the group with low prospects receive help that takes more time and are withheld from the labor market for $T_{delay}$ timesteps.

With different values for $k_{ij}$ we can now define different settings, which we call *intervention scenarios*. The difference between $k_{11}$and $k_{22}$ defines how different the effect of the intervention model is for the two different prospect groups, as intended by the PES. The differences between $k_{11}$ and $k_{21}$ and between $k_{12}$ and $k_{22}$ define how individuals are adversely affected if the prediction algorithm incorrectly classifies them. In this case, the groups receive the type of help that is intended for the other group.

The values for the different entries of $k_{ij}$ define how—in the abstract setting of our model—"attention" or "resources" are distributed across the different groups.

For better readability, the $k$-values presented in the text and the plots are multiplied with a factor of 500, so for the $k$-matrix for scenario 1 all entries will be displayed as 1. We will use the following scenarios:

1. *Balanced*: $k_{11} = k_{12} = k_{21} = k_{22} = 1$. This is the base scenario, where all prospect groups receive the same quantity and quality of help, which also has the same effect, independent of the actual labor market prospects.
2. *Onlylow*: $k_{11} = k_{21} = 1, k_{12} = k_{22} = 0$. Only individuals classified as low-prospect receive help. The effect of the help is independent of whether the classification is correct or not.
3. *Onlyhigh*: $k_{11} = k_{21} = 0, k_{12} = k_{22} = 1$ Only individuals classified as high-prospect receive help. The effect of the help is independent of whether the classification is correct or not.

4. *Balanced_errors_penalized*: $k_{11} = k_{22} = k_{12} = 1, k_{21} = 1/2$. Both high and low-prospect groups receive the same amount of help, but if the classification is incorrect, the help is only half as effective.

Detailed descriptions of the scenarios are given in Table A1 in the "Supporting information". The different scenarios show different targeting of the aid of the PES (no targeting in scenario 1, different ways of targeting in scenarios 2-4), and comparing them thus allows to answer what impact targeted aid vs. non-targeted aid has on the long-term fairness of public authority interventions.

**Spin up phase.** Each model run is started with a spin-up phase. The first 400 timesteps are run without the intervention model in order to allow the buildup of an initial historical dataset. The first 200 timesteps are discarded, and the remaining 200 are used as the initial historical dataset - the historical dataset that will be used in the first step of the model that includes the PES. With each further step of the model, the historical dataset will be extended. Thus, in the longer run, the historical set will more and more be influenced by the predictions made by the PES.

**Metrics.** In order to address trade-offs between different long-term fairness goals, we need to quantitatively define fairness for our setting. Our first metric is the Between Group skills Difference (*BGSD*), which is given by the difference of the mean skills between the two groups:

$$BGSD = \bar{s}_{real,\, x_{pr}=0} - \bar{s}_{real,\, x_{pr}=1} \tag{11}$$

This is a group fairness metric, and it is a property only of the data, not the prediction model. Our second metric is the fraction of the individuals that are predicted as low-prospect by the model but actually are high-prospect and would be classified as high-prospect by the model if the individuals would have the opposite protected attribute, which we call *counterfactual fraction*:

$$\frac{N\left(C_{p,x_{pr}} = low \wedge C_{p,x^*_{pr}} = high \wedge C_{tr} = high\right)}{N\left(C_p = low\right)} \tag{12}$$

Here, $N$ is the number of individuals as a function of the respective prospect groups: $C_p$ is the predicted prospect group, $C_{tr}$ is the true prospect group, and $x^*_{pr}$ is the opposite protected attribute of $x_{pr}$. Both metrics are computed at each timestep of the model. *BGSD* and *counterfactual fraction* represent different fairness goals, and a comparison is thus essential for addressing whether there are trade-offs between different long-term fairness goals.

As the third metric, we use Equal Opportunity, which is the difference in True Negative Rates (TNR) between the two groups:

$$EO = TNR_{priv} - TNR_{upriv} \tag{13}$$

where negative means predicted low prospect class. It is the fraction of low-prospect predictions that are really low prospect. Equal opportunity is a widely used fairness metric (e.g.[25,40]). Both counterfactual fraction and equal opportunity are properties not only of the data but also of the prediction model.

## Results

For each of the four scenarios, the model was run with the base prediction model and full prediction model, and with either the unbiased or biased labor market, resulting in four model combinations. In order to get a better feel and intuition for the model, we start by looking at a single model run in detail. Then, all scenarios and model configurations are compared with respect to the *BGSD* and *counterfactual fraction* metrics.

**Single model run (scenario: *onlylow*, model: *full prediction*).** A run with full prediction (including the protected attribute) and unbiased labor market for scenario *onlylow* is shown in Fig. 3. The prediction-model performance metrics are only available for the time the prediction model is active. Shown is part of the spin-up phase (the first 200 timesteps were discarded) and at timestep 400 the PES intervention model kicks in. This is clearly seen in many of the shown measures. The skills $s_{real}$ start to increase. For both the privileged and the underprivileged group, the mean skills increases, but it increases more for the underprivileged group, as can be seen by decreasing *BGSD*. At around timestep 500 the pool of job seekers reaches a new equilibrium in skills. The average time that the individuals, who found a job at a certain time-point were already unemployed ($T_u$), shows a more complex dynamic (panels in row two, which show $T_u$ and between group difference in $T_u$ (*BGTuD_current*). Shortly after the PES starts its intervention, $T_u$ increases for both groups and then decreases again. The fraction of underprivileged individuals in the current pool of job seekers (individuals looking for a job and individuals in the waiting group of the PES, *frac_upriv* in the plot) is on the order of 0.8. The background population has—per how we generate our population data—a fraction of underprivileged individuals of 0.5. The reason that it is higher in the active group is that individuals in the privileged group have on average higher skills, and are thus more likely to find a job soon, leading to this imbalance in the active group. The plot *fraction in waiting* shows the fraction of individuals from the privileged group and the fraction of individuals from the underprivileged group in the job-seeker pool that are currently in the waiting position (where the low prospect individuals receive help while being withheld from the labor market). In the beginning, this fluctuates strongly. This fluctuation stems from the fact that in the first timestep after the PES starts its intervention, all low-prospect individuals from the current pool are put in the waiting group, and the pool is then filled up with random individuals from the background population, which does not compensate all the low-skilled individuals. After around 100 timesteps this reaches an equilibrium and the strong fluctuations vanish. Clearly visible is that a much higher fraction of the underprivileged group ends up in the waiting group compared to the privileged group.

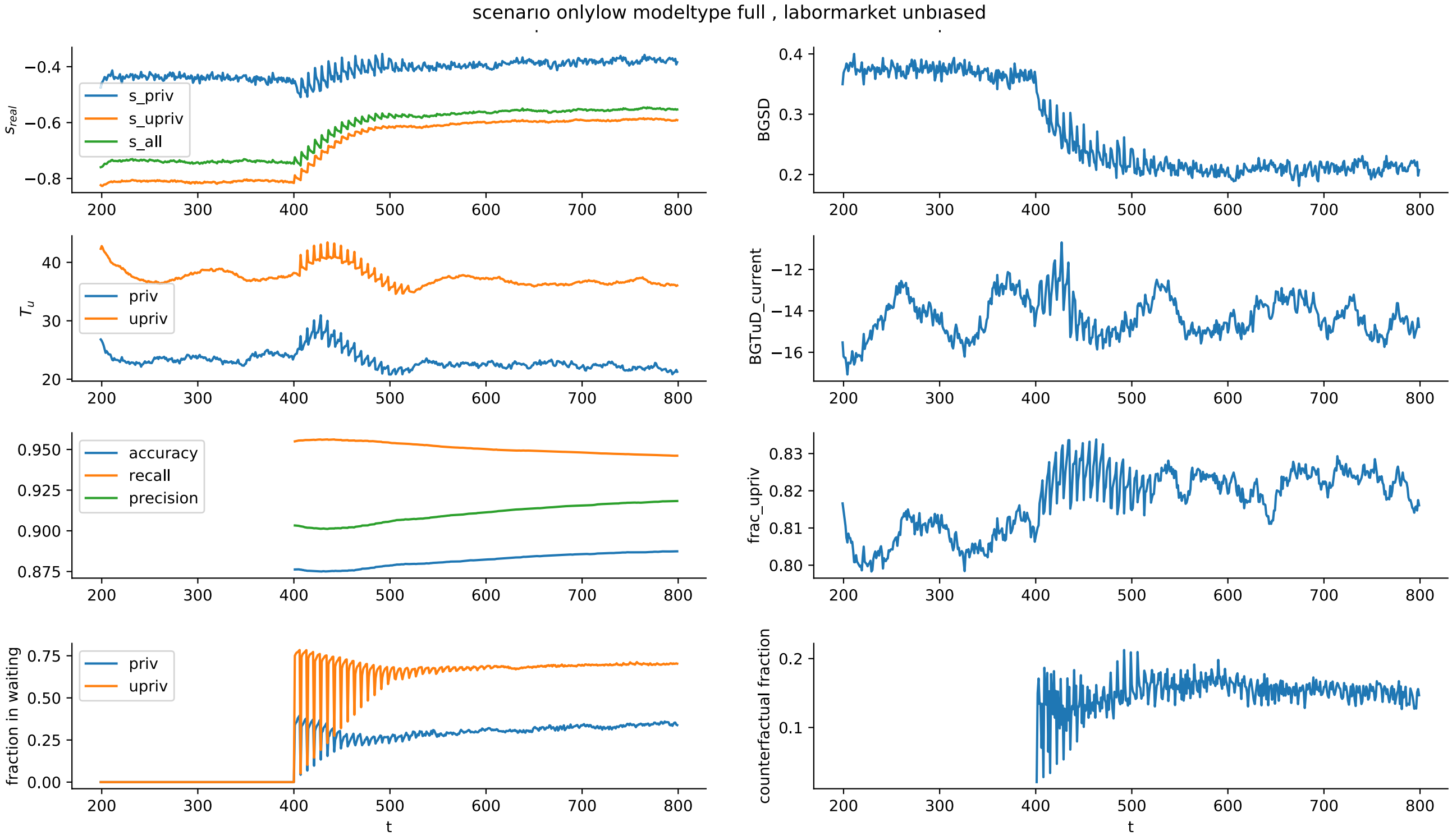

**Figure 3.** Time evolution of a single model run with the *onlylow* scenario and full (including protected attribute) prediction model. The First 200 timesteps are discarded, at timestep 400 the PES starts its intervention, which can be seen in several parameters (e.g. increase in $s_{real}$, decrease in *BGSD*). Some parameters/metrics are only available from the time the intervention starts.

**Comparison of scenarios: influence on group skill.** We now turn to a comparison of the different scenarios.

Figure 4 shows the time-evolution of *BGSD*, and Fig. 5 the average over the last 200 timesteps (a,b) and the difference between the average of the last and first 200 timesteps (c,d). The bars are split up by model type (full and base). In both figures, the left columns show the results for the runs with an unbiased labor market, and the right columns the results for the runs with a biased labor market. For the unbiased labor market, the PES with the base model has basically no effect on *BGSD*. With a biased labor market, the PES with base model *does* have an effect, but only for the intervention scenario in which only the high prospect group receives help. Interestingly, the *BGSD decreases* in this scenario. This is a relatively counter-intuitive result. Therefore, we inspect it in more detail. The full model evolution plots are shown in the SI (Fig. C1—C2). The skills of the privileged group actually decrease. This can be explained the following way: the high prospect group receives help, which will affect the privileged group more. So there is a large number with very high skills, and therefore near 1 probability of finding a job in the first timestep after entering the pool of job seekers. Additionally, the labor market is biased towards the privileged group, therefore also individuals with intermediate skills have a relatively large chance of finding a job immediately. Thus, for the pool of job seekers (both active and waiting) on which we measure *BGSD*, *BGSD* actually slightly decreases, as only the low-skilled ones from the privileged group remain in the pool. The result must also be connected with the fact that the individuals' low prospect group are put in the waiting group for $\Delta T_u = 5$ timesteps in the default model configuration, as when $\Delta T_u$ is set to zero, *BGSD* does not decrease in the high-only scenario (not shown).

If the PES uses the full prediction model (which includes the protected attribute as predictor), *BGSD* decreases for all scenarios, both in the unbiased and the biased labor market (Figs. 4c,d, 5c,d). The decrease in *BGSD* is smallest for the *onlyhigh* scenario. The other three scenarios have a very similar larger decrease in *BGSD*, which is larger in the biased labor market. This clearly shows that the targeting does affect the influence on between group differences. Another interesting result is that none of the scenarios reaches a *BGSD* of zero.

**Comparison of scenarios: fairness of prediction model.** We now turn to *counterfactual fraction* and equal opportunity, which measure different fairness goals than *BGSD*. In contrast to *BGSD* , they give insights into the fairness of the predictions model of the PES. The values for the end of the simulation are shown in Fig. 6. For the base model, *counterfactual fraction* is per definition zero, as the prediction model does not use the protected attribute, and changing it, therefore, has no effect. For the full model, on the other hand, changing the protected attribute has an effect. In the unbiased labor market, roughly 10-20% of the individuals classified as low prospect are in fact high prospect and are only classified as low prospect because of their protected attribute. In the biased labor market, the percentage is roughly twice as high. The latter effect is in our eyes hard to foresee intuitively without explicit modeling. In comparison to the results with respect to the *BGSD* metric clearly shows that there is a non-negligible trade-off between the two fairness goals. The full prediction model is better suited for decreasing *BGSD* (which is one fairness goal) in 3 of the 4 scenarios, but it introduces a non-zero *counterfactual fraction* (whose avoidance is another fairness goal). The second effect the full prediction model has is that it changes equal opportunity from slightly positive values (better for privilieged group) to negative values (worse for the privileged group—or more specifically, the true negative rate for the privileged group is smaller than for the underprivileged group).

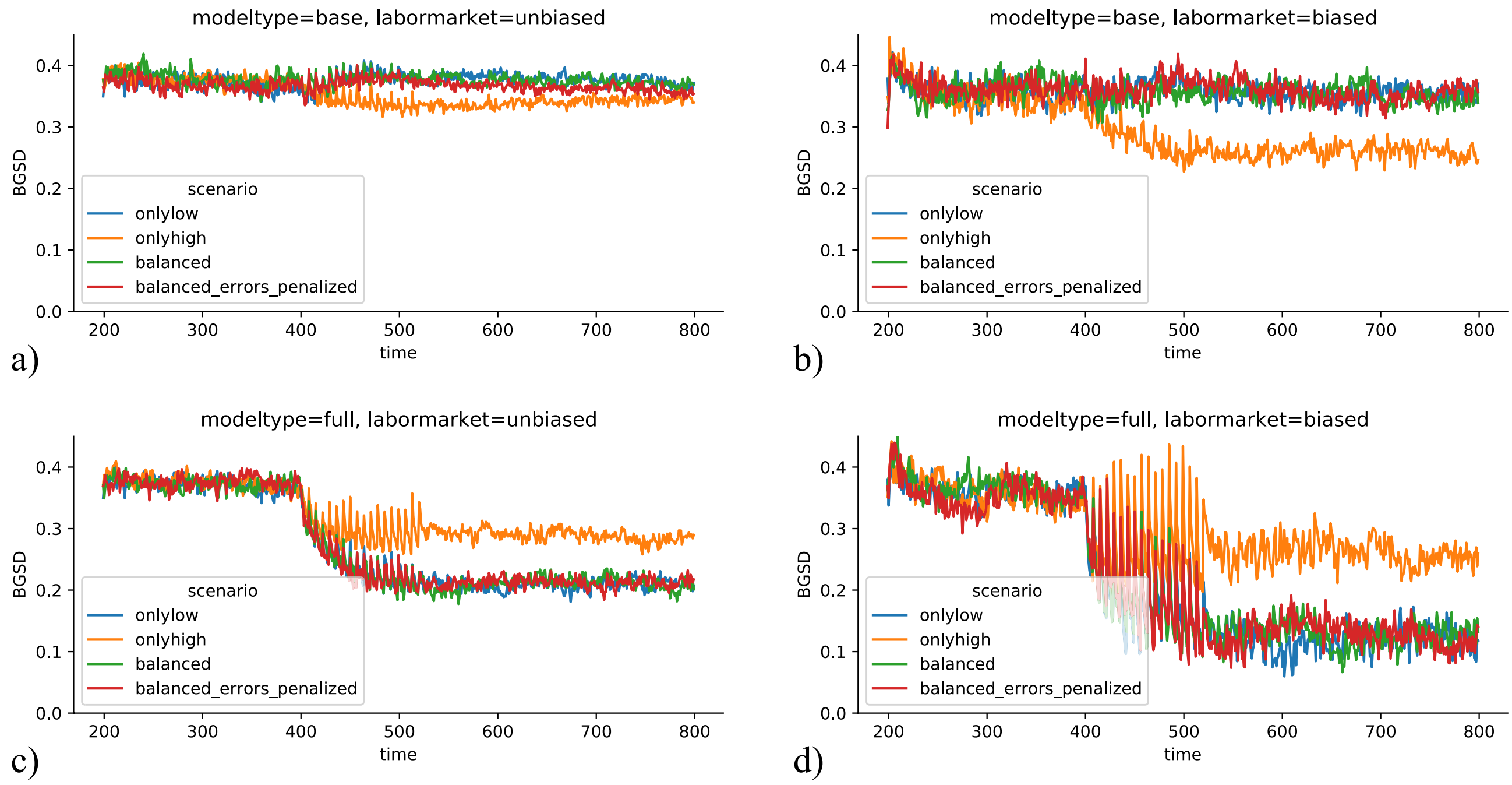

**Figure 4.** Time evolution of *BGSD* for all scenarios and all model combinations (full and base model, biased and unbiased prediction model). In an unbiased labor market and with the base prediction model, *BGSD* does not change significantly over time. Also, if the labor market is biased against the underprivileged group, *BGSD* does not change if the base prediction model is used, except if the help is targeted towards the high prospect group (scenario *onlyhigh*). If the full prediction model is used, *BGSD* decreases in all scenarios, independent of whether the labor market is biased or unbiased.

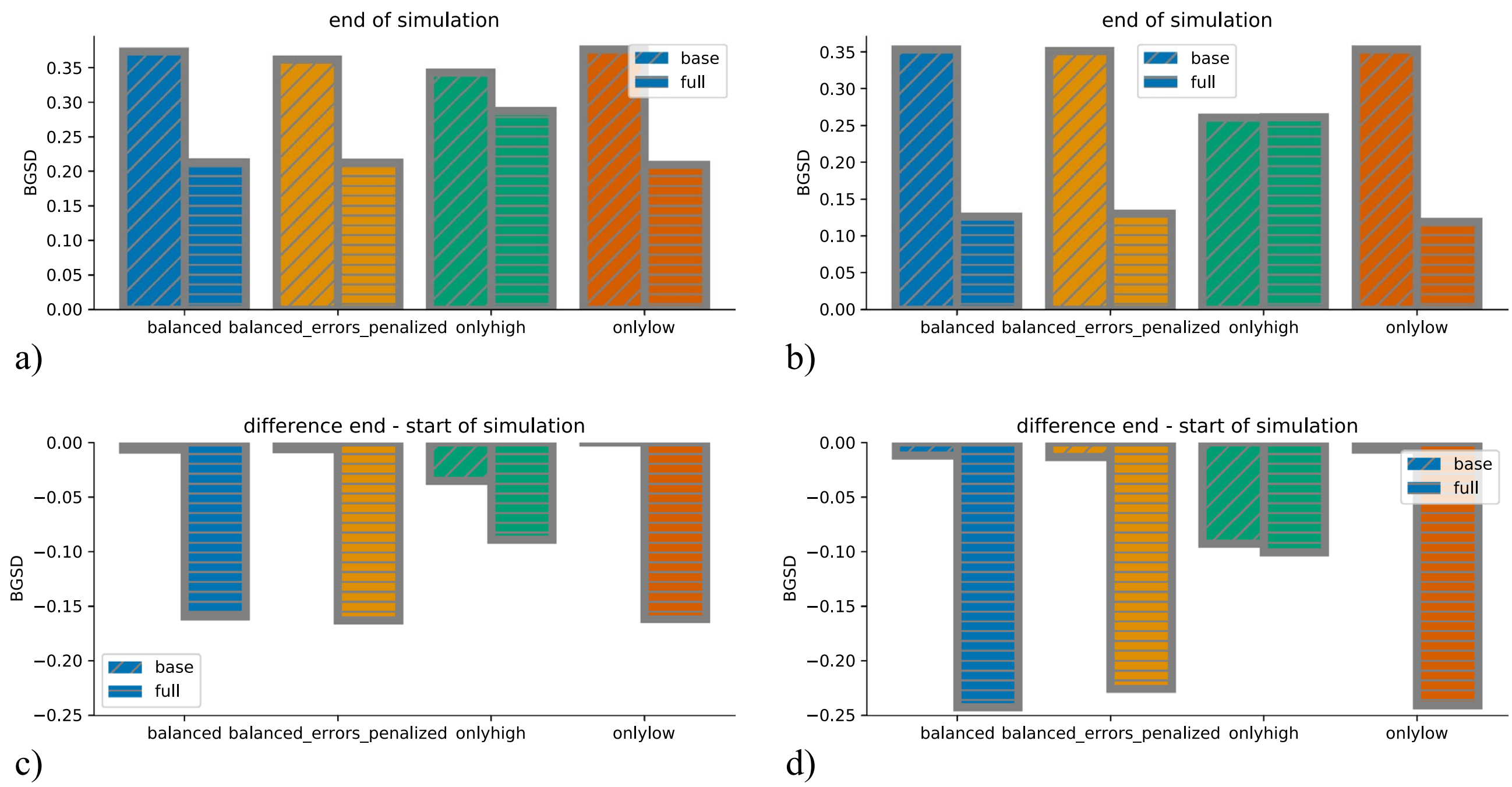

**Figure 5.** *BGSD* at the end of the simulations (**a**, **b**) and change of *BGSD* from start to end of simulations (**c**, **d**), for all intervention scenarios. a and c show the simulations for the unbiased, (**b** and **d**) the simulations for the biased labor market. Different colors indicate different intervention scenarios, and different hatching indicates the base (without protected attribute) and the full model (protected attribute).

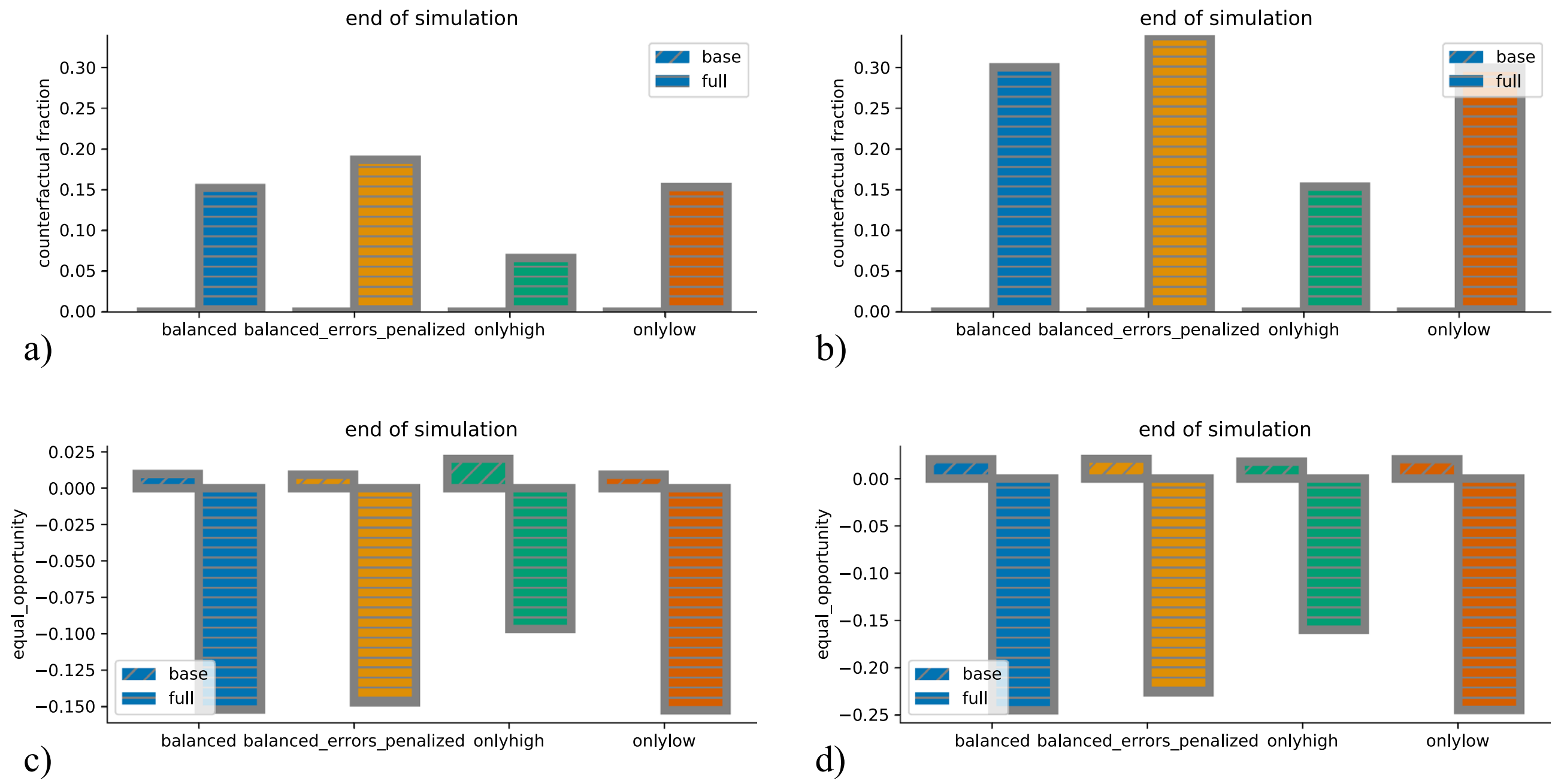

**Figure 6.** (**a**, **b**): Counterfactual fraction at the end of the simulations, for the unbiased (**a**) and biased (**b**) labor market. Different colors indicate different intervention scenarios, different hatching of the base (without protected attribute) and the full model (protected attribute). With the base prediction model, the *counterfactual fraction* is always zero. With the full prediction model, it is always positive, and depending on the intervention scenario. (**c**, **d**):Equal Opportunity ($TNR_{priv} - TNR_{upriv}$) at the end of the simulations, for the unbiased (**c**) and biased (**d**) labor market. Positive values indicate a positive bias in favor of the privileged group, and negative values a bias in favor of the unprivileged group.

## Discussion and limitations

In this study, we have taken a simplistic view of the labor market. Our personal skill model is just as complex as necessary to capture the main setting (skills unevenly distributed across two groups). Further, we assume that success in the labor market is based solely on individual skills and that observations of the labor market (e.g., who finds a job) are thus a measure of skills. In reality, this view has been challenged fundamentally, as luck is just as important for success as individual skills and abilities (see[41] and references therein). Additionally, personal factors such as sympathy can play a major role when selecting a candidate for an open position but will not be coded in the personal attributes potentially used by a PES (there is, however, the possibility that personal sympathy correlates with group-membership, thus further complicating things). Another important factor to consider is that both our models made the implicit assumption that there is, in principle, no shortage of jobs. If people have a high enough skill level, they will get a job (possibly). In our setting, there are enough jobs for the number of people, but not necessarily for their skill level. Real-world job-markets can, however, be limited on the side of open jobs, which would likely change the dynamics and effect of the PES's intervention.

While we included the option of having either a biased or unbiased labor market, the selection was fixed during the course of our simulations. In the real world, structural issues, such as biases in recruiters against or towards particular groups, may change over time. This raises the question of how representative—and, in the end, useful—historical recruitment data can be.

Also, in our abstract model-setting, the intervention scenarios we defined and applied in the study are only a subset of possible scenarios. In reality, even more different PES approaches would be conceivable, and this would need to be carefully reflected in the model setup.

Our intervention model (Eqs. (9) and (10) is deterministic. In reality, the effect of the intervention (increase in skill) will also have a random component, which one could, for example, model by adding noise. We did not do this in our study in order to keep the model as simple and comprehensible as possible.

This study is, therefore, a proof of concept and not an analysis of a real-world system. To study an existing real-world system (such as the Austrian AMS system), one would need to (i) have access to the data the PES uses—or at least to aggregated statistics—and (ii) carefully model the dynamics of the respective labor market. Here, the trade-off between complexity and completeness would need to be addressed in greater detail.

Despite the given limitations, we believe that our findings could inform the design of real-world systems reflecting labor markets. For example, we have shown that there exists a trade-off between reducing the disparity between a privileged and an unprivileged group and misclassifying individuals. This fundamental trade-off between group fairness and individual fairness reflects an important aspect that system designers need to take into account. One potential solution could be to focus on group fairness but implement additional measures to validate and mitigate potential misclassifications of individuals.

## Conclusion and future work

In this study, we investigated the long-term effects of data-driven intervention on the labor market in a simulated setting. Our results revealed an essential trade-off dilemma: the full model—in contrast to the base model—reduces *BGSD*, but at the same time classifies a number of individuals incorrectly as low-prospect solely because of their protected attribute, i.e., discriminates against them. Therefore, there is a trade-off between reducing the disparity between the two groups (reflected by a decrease in *BGSD*) and potentially treating individuals unfairly based on a protected attribute. Additionally, we found that active targeting of help (i.e., strategically distributing who receives what help) by the PES—compared to untargeted help—has little impact on inequality in the long-term, unless the help is targeted toward individuals with already high prospects, in which case inequality declines less. The purpose of this study was to show that in order to assess the ethical consequences of data-driven targeted support, e.g., for job-seekers, the investigation of long-term dynamics is crucial and requires careful quantitative modeling. This is not to say that other approaches (static quantitative approaches, philosophical/sociological approaches) are of less value, but that several perspectives are needed to give a complete picture. We have demonstrated this via a simple model for an employment market. Even in this relatively simple setting, it is not possible to answer questions on long-term fairness without explicit modeling.

A view on the long-term dynamics can be used for a more informed decision on whether to use a targeted support system. It can, however, also provide the basis for corrective actions that counteract unwanted long-term effects. Ideally, one would already consider long-term dynamics in the design phase. This is in line with[42] who propose the implementation of ethics by design rules, particularly in respect to biases, values, and the effect of modern technological development on individuals, and more general initiatives for ethically aligned design[43].

With clearly defined long-term goals and constraints and an accurate model for the long-term dynamics, data-driven targeted support systems could from the beginning on be designed in a way that prevents—or at least minimizes the risk of - unfair outcomes over all relevant timescales.

Future work should focus on enhancing and adopting the model to better reflect real-world situations of labor market interventions by Public Employment Services, e.g., by investigating settings with more than two real skill features and one protected feature. Furthermore, even in the setting, we studied here, there are a number of additional fairness-related questions that are worthy of being addressed. For example: what is the long-term effect of targeted help on general employment? What is the long-term effect on employment in each group? Are there trade-offs between the—ethically problematic—inclusion of protected attributes in the targeting versus the global goal of high employment?

In this study, the effects of prescribed intervention scenarios were studied. A different approach would be to reverse the problem and use reinforcement learning to find strategies that the PES can use for achieving certain goals.

Finally, the same approach—careful quantitative dynamical modeling—may be applied to other similar problems of distributing public resources, for example, in the context of education or public funding.

## Data availability

The software for this study was written in Python and is published in the first author's personal GitHub repository (https://github.com/sipposip/jobservice-ads-lterm-impact) and as FOSS via Zenodo (https://zenodo.org/record/6962331). It allows full reproduction of the results of this study as well as further experimentation with the model parameters.

## Acknowledgements

This work was supported by the "DDAI" COMET Module within the COMET – Competence Centers for Excellent Technologies Programme, funded by the Austrian Federal Ministry (BMK and BMDW), the Austrian Research Promotion Agency (FFG), the province of Styria (SFG) and partners from industry and academia. The COMET Programme is managed by FFG.

## Disclaimer
We have used a color scale for our graphics that at the first glance might seem suboptimal, as the colors are seemingly not very easy to distinguish. We have, on purpose, opted for that color scale, as it is, to our knowledge, the best color scale for that number of colors that still can also be interpreted by individuals with red-green color vision deficiency.

## Author contributions
S.S. conceived the idea for the study and developed the model. All authors contributed to the analysis, the interpretation of the results and the drafting of the manuscript.

## Competing interest
The authors declare no competing interests.

## Additional information
**Supplementary Information** The online version contains supplementary material available at https://doi.org/10.1038/s41598-023-28874-9.

**Correspondence** and requests for materials should be addressed to S.S.

**Reprints and permissions information** is available at www.nature.com/reprints.